\documentclass[conference]{IEEEtran}
\IEEEoverridecommandlockouts
\usepackage{graphicx}
\usepackage{subfigure}
\usepackage{amsfonts}
\usepackage{amsmath}
\usepackage{multirow,enumitem}
\usepackage{algorithmic}
\usepackage{algorithm}
\usepackage{xcolor}
\usepackage{bibnames}
\setlist[itemize]{leftmargin=*}
\newcommand{\m}{\textsc{MetaGAD}}

\def\BibTeX{{\rm B\kern-.05em{\sc i\kern-.025em b}\kern-.08em
    T\kern-.1667em\lower.7ex\hbox{E}\kern-.125emX}}
\begin{document}

\title{MetaGAD: Meta Representation Adaptation for Few-Shot Graph Anomaly Detection}

\author{
\IEEEauthorblockN{Xiongxiao Xu\IEEEauthorrefmark{1}, Kaize Ding\IEEEauthorrefmark{2}, Canyu Chen\IEEEauthorrefmark{1}, Kai Shu\IEEEauthorrefmark{3}}
\IEEEauthorblockA{\IEEEauthorrefmark{1}\textit{Department of Computer Science, Illinois Institute of Technology}, Chicago, IL, USA
\\\IEEEauthorrefmark{2}\textit{Department of Statistics and Data Science, Northwestern University}, Evanston, IL, USA 
\\\IEEEauthorrefmark{3}\textit{Department of Computer Science, Emory University}, Atlanta, GA, USA 
\\ xxu85@hawk.iit.edu, kaize.ding@northwestern.edu, cchen151@hawk.iit.edu, kai.shu@emory.edu}
}

\maketitle

\begin{abstract}
Graph anomaly detection has long been an important problem in various domains pertaining to information security such as financial fraud, social spam and network intrusion. The majority of existing methods are performed in an unsupervised manner, as labeled anomalies in a large scale are often too expensive to acquire. However, the identified anomalies may turn out to be uninteresting data instances due to the lack of prior knowledge. In real-world scenarios, it is often feasible to obtain limited labeled anomalies, which have great potential to advance graph anomaly detection. However, the work exploring limited labeled anomalies and a large amount of unlabeled nodes in graphs to detect anomalies is relatively limited. Therefore, in this paper, we study an important problem of few-shot graph anomaly detection. Nonetheless, it is challenging to fully leverage the information of few-shot anomalous nodes due to the irregularity of anomalies and the overfitting issue in the few-shot learning. To tackle the above challenges, we propose a novel meta-learning based framework, {\m}, that learns to adapt the knowledge from self-supervised learning to few-shot supervised learning for graph anomaly detection. In specific, we formulate the problem as a bi-level optimization, ensuring {\m} converging to minimizing the validation loss, thus enhancing the generalization capacity. The comprehensive experiments on six real-world datasets with synthetic anomalies and "organic" anomalies (available in the datasets) demonstrate the effectiveness of {\m} in detecting anomalies with few-shot anomalies. \textbf{The code is available at https://github.com/XiongxiaoXu/MetaGAD}.
\end{abstract}


\section{Introduction}
Graph structured data can be used to represent a large number of systems across various areas including social networks~\cite{garton1997studying}, physical systems~\cite{szklarczyk2015string}, knowledge graphs~\cite{nickel2015review}, neuroscience~\cite{behrouz2023admire++}, etc. However, real-world graphs can often be contaminated with a small portion of nodes, i.e., \textit{anomalies}, whose patterns deviate significantly from the majority of nodes. For example, in a social network that represents friendship relationships, there may exist camouflaged users who randomly follow other users to spread disinformation~\cite{dou2020enhancing}. In a cloud network, malicious nodes that produce abnormal traffic can significantly damage the resource management and weaken the security of the network environment~\cite{samrin2017review}. As the existence of even a few abnormal instances could cause extremely detrimental effects, it is imperative to design advanced graph anomaly detection algorithms to facilitate a secure and healthy network space. 

\begin{figure}[t!]
    \centering
    \subfigure[Raw representation]{
    	\begin{minipage}{0.23\textwidth}
   		 	\includegraphics[width=\textwidth]{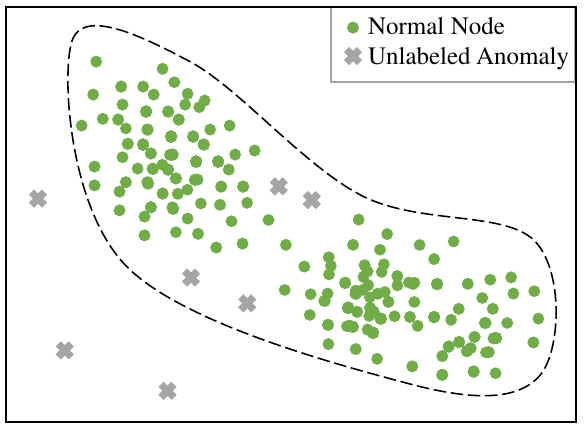}
    	\end{minipage}
    }
    \hspace{-0.3cm}
    \subfigure[Adapted representation]{
    	\begin{minipage}{0.23\textwidth}
   		 	\includegraphics[width=\textwidth]{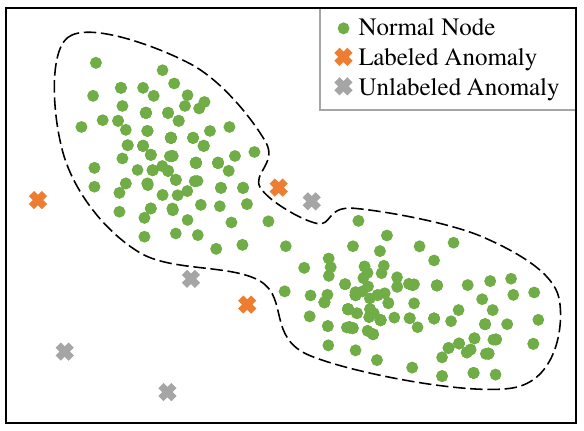}
    	\end{minipage}
    }
    \caption{An illustration of the representation gap: (a) the raw representation from a graph encoder, where the anomaly detector is limited in classifying anomalies; (b) while an adapted representation learned from the proposed model, instructed by the supervision of few-shot anomalies, can facilitate the anomaly detector to identify anomalies.}
    \label{fig:representation}
    \vspace{-0.6cm}
\end{figure}

Graph Neural Networks (GNNs)~\cite{kipf2016semi,wu2020comprehensive}, which generalize the deep neural networks to graph structured data, have shown promising representation learning capability on graphs and gained practical significance in graph anomaly detection~\cite{ma2021comprehensive,zhao2020error,ding2019deep}. For example, Ding \textit{et al.} propose a deep autoencoder to reconstruct both attributes and structures with graph convolutional networks (GCNs) to enhance detection performance~\cite{ding2019deep}. Chen \textit{et al.} utilize generative adversarial networks to inform anomaly evaluation measurement to make predictions~\cite{chen2020generative}. The majority of these methods are developed in an \textit{unsupervised} paradigm since it is usually labor-intensive and time-consuming for obtaining enough labels of anomalous nodes to train supervised models. However, unsupervised models may lead to sub-optimal anomaly detection results due to the complexity and diversity of anomalies and the lack of prior knowledge to guide the model training~\cite{ding2021few,liu2022bond}. In realistic scenarios, it is often feasible to obtain a limited amount of labeled anomalies from domain experts or user feedback~\cite{ding2019interactive}. Such labeled anomalies have great potential to boost the performance of graph anomaly detection~\cite{ding2021few}. Therefore, there is a pressing need to study the important problem of few-shot graph anomaly detection.

However, it is not trivial to detect anomalous nodes on a graph in a few-shot setting. 
First, it is challenging to find a principled way to leverage a large amount of unlabeled nodes and limited labeled anomalies. Recent work has shown that self-supervised learning~\cite{liu2022graph} can enhance graph learning tasks and anomalous nodes are regarded as auxiliary information in the self-supervised learning. However, such method focusing on unlabeled nodes may not fully utilize labeled anomalous nodes, resulting in sub-optimal performance. It remains unclear how to optimally focus on the primary task of learning from few-shot anomalies with the help of a large amount of unlabeled nodes. Second, the representation gap between self-supervised and supervised learning in graph anomaly detection may remain. For example, Figure~\ref{fig:representation} illustrates such representation gap across self-supervised and supervised learning paradigms. It shows that the raw representation directly derived from a graph encoder is not be appropriate for an anomaly detector to distinguish anomalous nodes from normal nodes. How to mitigate the representation gap to enhance graph anomaly detection performance remains unclear. Third, the overfitting issue is notoriously common in a few-shot learning setting. The performance of most of the existing methods is inevitably undermined by the overfitting issue under the few-shot setting. For example, when training with few labeled nodes with a pretrain-finetune approach, an over-parametric deep GNN model tends to overfit~\cite{ding2022meta}. A novel learning algorithm that is able to fully leverage the information of few-shot anomalies while alleviating the overfitting issue is highly desirable.

To address the aforementioned challenges, we propose {\m}, a meta-learning based framework, that \textit{learns to adapt the node representations learned from self-supervised learning to maximally facilitate few-shot supervised learning}. {\m} is built upon a Representation Adaptation Network (RAN) and an anomaly detector, which are connected by a novel meta-learning algorithm. In particular, RAN, as the meta-learner, learns to adapt the node representation from self-supervised learning for few-shot supervised learning. By incorporating the feedback reflecting generalization performance, RAN can learn to dynamically adapt the raw representation to the anomaly-aware representation. The anomaly detector, as the target model, can precisely separate the anomalies and normal nodes based on the adapted representation. Upon the proposed meta-learning algorithm, RAN and the anomaly detector are able to enhance each other synergistically where a bi-level optimization guarantees the convergence is towards the decrease of the validation loss, thus avoiding the overfitting issue. To summarize, the main contributions are as follows:

\begin{itemize}
    \item We study an crucial and practical problem of few-shot graph anomaly detection;
    \item We propose a novel meta-learning approach {\m} that learns to adapt node representations from a self-supervised learning to maximally facilitate a supervised learning with few-shot anomalies. With the meta-leaning framework, the few-shot graph anomaly detection task is formulated as a bi-level optimization problem, which ensures that {\m} avoids the overfitting issue and leads to the enhanced generalization ability.
    \item We conduct extensive experiments on six real-world datasets with synthetic anomalies and real anomalies. The experimental results demonstrate the superiority of the proposed framework MetaGAD. 
\end{itemize}

\section{Related Work}\label{sec:related}
In this section, we describe the related work on graph anomaly detection and few-shot graph learning.

\subsection{Graph Anomaly Detection.}
Graph anomaly detection problem have a specific focus on the network structured data. Earlier research mostly study the problem in an unsupervised manner~\cite{liu2022pygod,zheng2021generative,roy2023gad,zhang2022reconstruction,duan2023graph,zhang2022reconstruction,xiao2023counterfactual}. However, the detection performance may be limited due to the lack of prior knowledge on the anomalies. Recently, researchers also propose to use GNNs for graph anomaly detection due to their strong modeling capacity~\cite{liu2021anomaly}. DOMINANT~\cite{ding2019deep} achieves superior performance over other shallow methods by building a deep autoencoder architecture on top of the graph convolutional networks. GAAN~\cite{chen2020generative} is a generative adversarial graph anomaly detection framework where a generator can fake graph nodes with Gaussian noise as input, and a discriminator is trained to identify whether two connected nodes are from the real graph or fake graph. The aforementioned approaches are designed in an unsupervised manner, which can not be directly applied to the scenarios where labeled anomalies are few. Some work focuses on graph anomaly detection with limited labels. For example, Meta-GDN~\cite{ding2021few} considers the few-shot graph anomaly detection problem with multiple graphs and it utilizes meta-learning to transfer from multiple auxiliary graphs to a target graph. Semi-GNN~\cite{wang2019semi} focus on the financial fraud detection problem with the multi-view labeled and unlabeled data. SAD~\cite{tian2023sad} employs a temporal graph network and an anomaly detector with a time-equipped
memory bank to detectt anomaly over dynamic graph. AugAN~\cite{zhou2023improving} is a data augmentation method to boost GAD model generalizability by enlarging training data and a customized training strategy. Other approach treats few-shot labeled anomalies as additional auxiliary information to assist the main-task of self-supervised learning to detect anomalies~\cite{zheng2022unsupervised}. Different from these work, we study an important and practical few-shot anomaly detection problem in a single homogeneous graph. We propose a meta-learning based framework {\m} to learn to adapt the node representations from self-supervised learning to maximally assisting main-task of few-shot supervised learning.

\subsection{Few-Shot Graph Learning.}
Graph neural networks have shown great success for modeling graph data with neural networks~\cite{wu2020comprehensive,kipf2016semi,zhou2020graph}. The majority of existing graph learning algorithms focus on the scenario where enough labeled instances are available for training, which is often infeasible as collecting such auxiliary knowledge is laborious and requires intensive domain-knowledge~\cite{yao2020graph,ding2020graph}. Recently, few-shot graph learning is attracting increasing attention for graph learning under limited labeled data, which can be generally categorized into two types~\cite{ding2022graph}: (1) gradient-based methods, which aim to learn a better initialization of model parameters that can be updated by a few gradient steps in future tasks~\cite{ding2022graph,zhou2019meta,liu2021relative}; and (2) metric-based approaches, which propose to learn a generalized metric and matching functions from training tasks~\cite{ding2020graph,guo2021few}. For example, Liu \textit{et al.} utilize a hub-based relative and absolute location embedding with meta-learning for few-shot node classification~\cite{liu2021relative}. Yao \textit{et al.} attempt to incorporate prior knowledge learned from auxiliary graphs by constructing prototypes in hierarchical graph neural networks~\cite{yao2020graph}. In addition, Ding \textit{et al.} propose a graph prototypical network by considering the informativeness of labeled nodes in a meta-learning technique for few-shot node classification~\cite{ding2020graph}. 

\section{Problem Definition}\label{sec:problem}
We first introduce the notations of this paper, and then give the formal problem definition. Let $G=(\mathcal{V}, \mathcal{E}, \mathbf{X})$ denote a graph where $\mathcal{V}=\{v_1,\cdots,v_n\}$ is the set of $n$ nodes, and $\mathcal{E}=\{e_1,\cdots,e_m\}$ denotes the set of $m$ edges, and $\mathbf{X}=[\mathbf{x}_1;\cdots; \mathbf{x}_n]\in\mathbb{R}^{n\times d}$ is the feature matrix for all nodes where $\mathbf{x}_i$ is the feature vector for node $v_i$. The graph can also be denoted as $G=(\mathbf{A},\mathbf{X})$ where $\mathbf{A}\in\{0,1\}^{n \times n}$ denotes the adjacency matrix, where $\mathbf{A}_{ij}=1$ indicates there is an edge between node $v_i$ and $v_j$; otherwise $\mathbf{A}_{ij}=0$. 

The goal of few-shot graph anomaly detection is to incorporate the knowledge of few-shot labeled anomalies in addition to exploiting the attribute and structure information of the graph. Let $\mathcal{V}_l$ denote the few-shot labeled anomalies and $\mathcal{V}_u$ denote the unlabeled nodes. Note that $\mathcal{V}=\{\mathcal{V}_l, \mathcal{V}_u\}$ and $|\mathcal{V}_l|\ll|\mathcal{V}_u|$, since only limited nodes are labeled as anomalies. Following the common setting, we treat anomaly detection problem as a ranking problem~\cite{ma2021comprehensive} and formally define our problem as follows:
\begin{center}
\fbox{\parbox[c]{.95\linewidth}{\textbf{Problem Statement:}
Given $G=(\mathbf{A},\mathbf{X})$ with few-shot labeled anomalies $\mathcal{V}_l$, we aim to learn an anomaly detection model that can detect anomalies effectively on the graph $G$. Ideally, the model can assign higher ranking scores to the anomalies than that of the normal nodes.
}}
\end{center}

\begin{figure*}[!t]
    \centering
    \includegraphics[width=\textwidth]{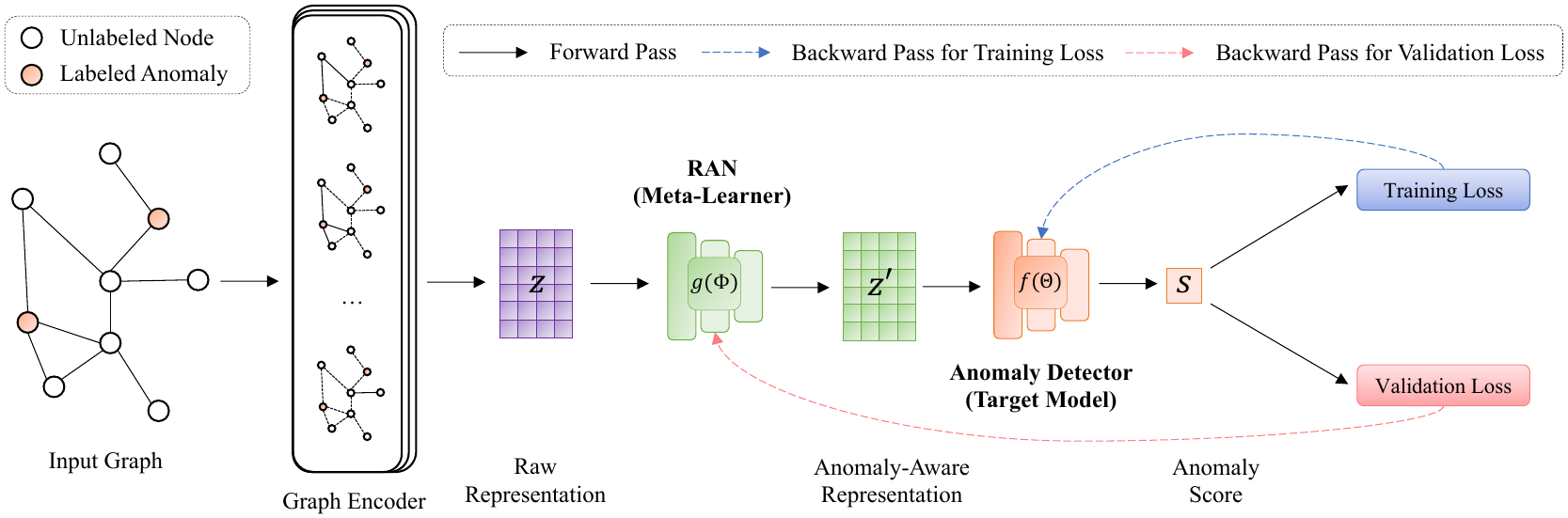}
     \caption{The illustration of {\m}. The anomaly detector (target model) parameter $\Theta$ is optimized by the training loss and RAN (meta-learner) parameter $\Phi$ is optimized by the validation loss. Such meta-learning algorithm enables RAN and the anomaly detector enhance each other synergistically.
     }
     \vspace{-0.3cm}
    \label{fig:framework}
\end{figure*}

\section{Methodology}\label{sec:methodology}
To solve the few-shot graph anomaly detection, we propose a new framework {\m} as shown in the Figure~\ref{fig:framework}. {\m} learns to adapt the node representations from self-supervised learning to maximally facilitate few-shot supervised learning. Upon the proposed meta-learning algorithm, the meta-learner, Representation Adaptation Network (RAN), learns to adapt the raw representations to the anomaly-aware representation and largely overcomes the overfitting issue, using the feedback from validation loss. The target model, anomaly detector, can precisely detect anomalous nodes based on the adapted representation.

\subsection{The Framework of {\m}.}\label{sec:framework}
In order to leverage a large number of unlabeled nodes to facilitate the supervised learning with few-shot anomalies, we propose a framework {\m} including three major components: (1) a graph encoder, (2) a representation adaptation network, and (3) an anomaly detector.
\\\textbf{Graph Encoder.} To learn effective node representation from the graph, we utilize Graph Neural Networks (GNNs) to extract structure and feature information of the graph. In general, GNNs have a message-passing mechanism to aggregate attributes from local neighbourhood, and they can transform high-dimension features to low-dimension network embedding. The multi-layer GNNs can be stacked as a graph encoder to further strengthen capability. Formally, the graph encoder is described as:
\begin{equation*}
\mathbf{Z} = Encoder(\mathbf{A}, \mathbf{X})
\end{equation*}
Here, $\mathbf{Z}$ is the output (the last layer of GNNs) of a graph encoder and $Encoder$ denotes the graph encoder.
There are some pretrained graph encoders we can use from the existing self-supervised graph anomaly detection methods, including generative models~\cite{ding2019deep,chen2020generative} and contrastive models~\cite{liu2021anomaly,xu2022contrastive}. We choose the encoder from a representative generative model DOMINANT~\cite{ding2019deep} as the graph encoder because the generative models better characterize original data distribution~\cite{liu2021self}. 
\\\textbf{Representation Adaptation Network (Meta-Learner).}
While representation $\mathbf{Z}$ implies knowledge of the graph, it is not instructed by any supervision of anomalous nodes. To overcome the issue in Figure~\ref{fig:representation}, we propose a Representation Adaptation Network (RAN) to learn to adapt the raw representation $\mathbf{Z}$ to the anomaly-aware representation $\mathbf{Z}'$. Specifically, we define a RAN as $g_\Phi = \mathbb{R}^d\rightarrow{}\mathbb{R}^d$, where $d$ is the dimension of node representation. Conceptually, any network with proper input and output sizes is feasible. We employ a two-layer feed-forward network because such simple architecture can avoid heavy parameter overhead. Furthermore, the two-layer feed-forward network has been empirically validated work well to adapt the representation in the text tasks~\cite{pfeiffer2020adapterhub,pfeiffer2020mad}. We utilize the RAN on top of the raw representation $\mathbf{Z}$ as follows:
\begin{equation*}
\mathbf{Z'} = g_\Phi(\mathbf{Z}) = \mathbf{W}_2ReLU(\mathbf{W}_1\mathbf{Z} + \mathbf{b}_1)+\mathbf{b}_2
\end{equation*}
where $\Phi = \{\mathbf{W}_1, \mathbf{W}_2, \mathbf{b}_1, \mathbf{b}_2\}$ is the set of trainable parameters of RAN and $ReLU$ is the activation function. During the training, RAN is updated by the validation loss. The supervision information of anomalous nodes in the validation loss makes RAN capable of adapting the raw representation to the anomaly-aware representation. Furthermore, it can largely alleviate the overfitting issue. The role of the validation loss is like the reward in reinforcement learning~\cite{sutton2018reinforcement} somehow as it can provide feedback of generalization performance of {\m}. The feedback can be incorporated by RAN and guides the RAN to update its parameters in real-time, thus largely alleviating the overfitting issue.
\\\textbf{Anomaly Detector (Target Model).}
Based on the  anomaly-aware representation from RAN, an anomaly detector is able to precisely estimate the abnormality of a node $v_i \in \mathcal{V}$:
\begin{equation*}
s_i = f_\Theta(\mathbf{Z}'_{i,:})
\end{equation*}
where $f_\Theta(\cdot)$ is a multi-layer perceptron (MLP), $s_i$ is the anomaly score of the node $v_i$. During the training, the anomaly detector is updated by the training loss because it requires more data to train. In general, RAN and the anomaly detector are updated by the validation loss and training loss, respectively. Such novel design is appropriate for the desired training data size. The anomaly detector is randomly initialized and lack any prior information, and thus the anomaly detector needs a substantial amount of data. However, the raw representation from the graph encoder has contained structure and feature information of the graph, albeit not perfect enough for the few-shot anomaly detection. Consequently, RAN only requires relatively less data to perform effectively. 
\\\textbf{Cost-Sensitive Loss Function.}
After we evaluate an anomaly score for each node, we calculate the loss function to train the model. We regard the labeled anomalies $\mathcal{V}_l$ as positive instances and the unlabeled nodes $\mathcal{V}_u$ as negative instances. Specifically, the label of node $v_i$ is 1 (i.e., $y_i=1$) if the node $v_i$ is labeled anomaly (i.e., $v_i\in\mathcal{V}_l$); the label of node $v_i$ is 0 (i.e., $y_i=0$) if the node $v_i$ is unlabeled nodes (i.e., $v_i\in\mathcal{V}_u$). Since the number of positive instances is much less than that of negative instances, i.e., $|\mathcal{V}_l|\ll|\mathcal{V}_u|$, we propose a cost-sensitive learning strategy~\cite{elkan2001foundations,cao2019learning,cui2019class} into the few-shot graph anomaly detection to investigate the impact of the class imbalance. In detail, we propose a cost weight hyperparameter $w$ to adjust the weight of positive instances in the training. Thus, the cost-sensitive loss function can be written as follows:
\begin{align*}\label{eq:5}
& L = -\frac{1}{n}\sum_{i=1}^{n}[w * y_i\log(\sigma(s_i))+(1-y_i)\log(1-\sigma(s_i))]
\end{align*}
where $L$ is the loss function, and $w$ is the cost weight to balance between positive instances and negative instances, and $\sigma$ is the sigmoid function. We discuss the impact of $w$ to the detection performance in the Section~\ref{sec:imbalance}. We find that it is not necessary to make positive instances and negative instances balanced; instead, the skewed state in a degree improve the detection performance. We attribute the interesting observation to the nature of the anomaly detection problem.

\subsection{A Meta-Learning Approach.}\label{sec:meta-learning}
To enable RAN (\textit{meta-learner}) and the anomaly detector (\textit{target model}) enhance each other synergistically, we propose a novel meta-learning method for them.
\\\textbf{Learning to Adapt Representation.}
The goal of RAN is to adapt representation and the goal of the anomaly detector is to precisely distinguish anomalies from normal nodes based on the adapted representation. To connect RAN and the anomaly detector in a principled way, we propose a meta-learning framework where the meta-learning objective is as follows: \textit{if RAN $g_\Phi$ can effectively adapt the representation, such adapted representation $\mathbf{Z}'=g_{\Phi}(\mathbf{Z})$ should be more beneficial to the anomaly detector than the raw representations $\mathbf{Z}$, such that the model achieves a smaller validation loss and largely alleviate the overfitting issue.}
\\\textbf{Bi-level Optimization.}
Formally, the above meta-objective implies a bi-level optimization as follows:
\begin{align*}
&\min_{\Phi}L_{val}(\Theta^*(\Phi),\Phi)\nonumber\\
&{s.t.}\;\Theta^*(\Phi) = \arg\min_{\Theta}L_{train}(\Theta,\Phi)
\end{align*}
where $L_{train}$ and $L_{val}$ denote the training loss and the validation loss, respectively. The above bi-level optimization function prioritizes optimizing the upper objective function, i.e., minimizing the validation loss, rather than the lower objective function, i.e., minimizing the training loss. Particularly, when the training loss and the validation loss do not decrease at the same time, i.e., the training loss decreases but the validation loss increases, also known as \textit{notorious} overfitting problem~\cite{tetko1995neural}, the bi-level structure guarantees the optimization direction is towards minimizing the validation loss. Therefore, it can effectively alleviate the overfitting problem and promote the generalization performance.
\\\textbf{An Approximate Optimization Algorithm.}
Analytic solution for the above bi-level optimization is computationally infeasible as it requires solving for the optimal $\Theta^*$ whenever $\Phi$ gets updated. For ease of computation, we propose to alternatively update parameters with one-step stochastic gradient descent (SGD) to approximate the analytic solution~\cite{liu2018darts,xia2021metaxl}: 

\begin{algorithm}[tbp!]
\caption{The optimization process for {\m}}\label{alg}
\begin{algorithmic}[1]
\STATE Initialize randomly the anomaly detector parameters $\Theta$ and RAN parameters $\Phi$.
\WHILE{not converged}
\STATE \textit{\# Target Model Update}
\STATE $\Theta: \Theta'=\Theta-\alpha\nabla_{\Theta}L_{train}(\Theta,\Phi)$
\STATE \textit{\# Meta-Learner Update}
\STATE  $\Phi' = \Phi - \beta\nabla_{\Phi}L_{val}(\Theta - \alpha\nabla_{\Theta}L_{train}(\Theta,\Phi),\Phi)$
\ENDWHILE
\end{algorithmic}
\end{algorithm}

\textbf{Target Model (Anomaly Detector) Update:}
\begin{equation}\label{eq:1}
    \Theta' = \Theta - \alpha\nabla_{\Theta}L_{train}(\Theta,\Phi)
\end{equation}
where $\Theta'$ denotes the updated $\Theta$ after a one-step SGD and $\alpha$ is the corresponding learning rate. 

\textbf{Meta-Learner (RAN) Update:}
\begin{equation}\label{eq:2}
    \Phi' = \Phi - \beta\nabla_{\Phi}L_{val}(\Theta - \alpha\nabla_{\Theta}L_{train}(\Theta,\Phi),\Phi)
\end{equation}
where $\Phi'$ denotes the updated $\Phi$ after a one-step SGD and $\beta$ is the corresponding learning rate.

We expand the gradient of Equation~\ref{eq:2} by applying the chain rule as follows:
\begin{align}\label{eq:3}
    &\nabla_{\Phi}L_{val}(\Theta - \alpha\nabla_{\Theta}L_{train}(\Theta,\Phi),\Phi)\nonumber\\
    &=\nabla_{\Phi}L_{val}(\Theta',\Phi)-\alpha\nabla_{\Phi,\Theta}^2L_{train}(\Theta,\Phi)\nabla_{\Theta'}L_{val}(\Theta', \Phi)
\end{align}
The second term in Equation~\ref{eq:3}, which involves an expensive matrix-vector product, can be approximated for light computation with the technique of finite difference approximation as follows:
\begin{align}
    &\nabla_{\Phi,\Theta}^2L_{train}(\Theta,\Phi)\nabla_{\Theta'}L_{val}(\Theta', \Phi))\nonumber\\
    &\approx\frac{\nabla_{\Phi}L_{train}(\Theta^+,\Phi)-\nabla_{\Phi}L_{train}(\Theta^-,\Phi)}{2\epsilon}
\end{align}
where $\Theta^\pm=\Theta\pm\epsilon\nabla_{\Theta'}L_{val}(\Theta',\Phi)$ and $\epsilon$ is a small constant.

During training, we alternatively update $\Theta$ and $\Phi$ following Equation~\ref{eq:1} and Equation~\ref{eq:2} until convergence, respectively, as shown in Algorithm~\ref{alg}.

\section{Experiments}\label{sec:exp}
In this section, we conduct experiments on datasets with synthetic anomalies and real anomalies to demonstrate the effectiveness of the {\m}. Specifically, we aim to answer the following research questions (RQs): 
\begin{itemize}
    \item \textbf{RQ1:} How effective is {\m} for detecting anomalies with few or even one labeled instance?
    \item \textbf{RQ2:} How does each key module of {\m} contribute to the detecting performance?
    \item \textbf{RQ3:} How does {\m} alleviate the commonly overfitting issue in the few-shot problem?
    \item \textbf{RQ4:} How does the performance of {\m} change under different imbalance levels in the data?
    \item \textbf{RQ5:} How robust is the {\m} under different contamination levels in the unlabeled nodes?
\end{itemize}

\subsection{Datasets.} \label{sec:dataset}
To comprehensively evaluate the {\m} model, we adopt six widely-used real-world datasets that have been used in previous research~\cite{liu2021anomaly, liu2022bond}. The statistics of the datasets are shown in Table~\ref{tab:dataset}, and the detailed descriptions are as follows:
\begin{itemize}
    \item \textbf{Cora}, \textbf{Citeseer} and \textbf{Amazaon Photo} are three citation network and social network datasets. There are no ground-truth anomalies given in the three datasets.
    \item \textbf{Wiki}, \textbf{Amazon Review} and \textbf{YelpChi} are three editor-page and review datasets with organic anomalies. The ground-truth anomalies are available in the three datasets.
\end{itemize}

As there are no ground-truth anomalies given in the Cora, Citeseer and Amazon Photo datasets, we adopt two well-recognized methods to synthesize anomalies following previous research~\cite{liu2022bond, ding2021few, ding2019deep, liu2021anomaly}. The details are as follows:
\begin{itemize}
    \item \textbf{Structural Anomaly} are anomalous nodes related to their edges. They are obtained by perturbing the topological structure of the attributed network. The structure of chosen nodes are changed by adding more links in attributed network. The intuition is that a clique where a set of nodes are fully connected each other is regraded as a typical abnormal structure in many real-world scenarios~\cite{hooi2016fraudar,skillicorn2007detecting}. For example, members of fraud organizations are closely connected to each other to facilitate the crime. To be more specific, we randomly pick $m$ nodes, and make them fully connected each other. Then we repeat this process for $n$ times. Afterwards, we get $m*n$ structural anomaly where there are $n$ cliques and each clique has $m$ nodes.
    \item \textbf{Contextual Anomaly} are anomalous nodes regarding to their features. They are acquired by perturbing the nodal features in the attributed network. The feature of a selected node is replaced by that of a "sufficiently remote" node. The intuition is that node is similar to its neighbourhood. For example, people in the same community are more likely to have same hobbies~\cite{yuan2021higher}. Specifically, for each randomly picked node $i$, we randomly select $k$ nodes other than node $i$ in attributed network, and compute the distance between feature of node $i$ and each node $j$ in the selected $k$ nodes, e.g., $||x_i-x_j ||_2$. Then we find the node $j$ with the maximum distance in the selected $k$ nodes. Afterwards, we replace the feature of node $i$ with that of node $j$.
\end{itemize}
Following the existing works~\cite{liu2022bond, liu2021anomaly}., we synthesize equal number of structural anomalies and contextual anomalies

\begin{table}[!t]
    \centering
    \caption{The statistics of three datasets with synthetic anomalies (indicated by *) and three datasets with real anomalies.}
    \begin{tabular}{l|cccccc}
    \hline
        \textbf{Dataset} & \textbf{\#Nodes} & \textbf{\#Edges} & \textbf{\#Attr.} & \textbf{\#Anomaly} & \textbf{Ratio}\\ \hline
        \textbf{Cora*} & 2,708 & 11,606 & 1,433 & 150 & 5.54\% \\
        \textbf{Citeseer*} & 3,327 & 10,154 & 3,703 & 150 & 4.51\% \\ 
        \textbf{Amazon Photo*} & 7,650 & 146,810 & 745 & 450 & 5.90\% \\ 
        \textbf{Wiki} & 9,227 & 18,257 & 64 & 217 & 2.35\% \\ 
        \textbf{Amazon Review} & 11,944 & 8,796,784 & 25 & 821 & 6.87\% \\ 
        \textbf{YelpChi} & 23,831 & 98,630 & 32 & 1,217 & 5.11\% \\
    \hline
    \end{tabular}\label{tab:dataset}
\end{table}

\begin{table*}[!t]
    \centering
    \caption{Performance comparison results on the datasets with synthetic anomalies. Best results are shown in bold.}
    \begin{tabular}{l|cc|cc|cc}
    \hline
        \multirow{2}{*}{\textbf{Methods}} & \multicolumn{2}{c|}{\textbf{Cora}} & \multicolumn{2}{c|}{\textbf{Citeseer}}&\multicolumn{2}{c}{\textbf{Amazon Photo}} \\ \cline{2-7}
        & AUC-ROC & AUC-PR & AUC-ROC & AUC-PR & AUC-ROC & AUC-PR\\ \hline \hline
        \textbf{Radar} & 0.4349$\pm$0.0000 & 0.0580$\pm$0.0000 & 0.4895$\pm$0.0000 & 0.0407$\pm$0.0000 & 0.5394$\pm$0.0000 & 0.0627$\pm$0.0000 \\
        \textbf{ANOMALOUS} & 0.4313$\pm$0.0005 & 0.0577$\pm$0.0001 & 0.4857$\pm$0.0000 & 0.0404$\pm$0.0000 & 0.4777$\pm$0.0004 & 0.0562$\pm$0.0000 \\ \hline
        \textbf{GAAN} & 0.5174$\pm$0.0416 & 0.0639$\pm$0.0051 & 0.5799$\pm$0.0415 & 0.0842$\pm$0.0390 & 0.5888$\pm$0.0041 & 0.1604$\pm$0.0140 \\
        \textbf{OCGNN} & 0.7950$\pm$0.0093 & 0.1942$\pm$0.0118 & 0.8050$\pm$0.0037 & 0.1449$\pm$0.0095 & 0.3664$\pm$0.0081 & 0.0421$\pm$0.0005 \\
        \textbf{DOMINANT} & 0.7649$\pm$0.0067 & 0.2994$\pm$0.0043 & 0.8375$\pm$0.0025 & 0.2579$\pm$0.0054 & 0.7119$\pm$0.0346 & 0.1003$\pm$0.0080 \\
        \textbf{CoLA} & 0.7696$\pm$0.0648 & 0.2028$\pm$0.0839 & 0.7812$\pm$0.0183 & 0.1816$\pm$0.0556 & 0.5519$\pm$0.0597 & 0.0740$\pm$0.0130 \\
        \textbf{ANEMONE} & 0.7702$\pm$0.0158 & 0.1451$\pm$0.0374 & 0.7429$\pm$0.0367 & 0.1480$\pm$0.0637 & 0.6688$\pm$0.0162 & 0.1574$\pm$0.0022 \\
        \textbf{CONAD} & 0.7658$\pm$0.0047 & 0.3004$\pm$0.0058 & 0.8378$\pm$0.0027 & 0.2579$\pm$0.0055 & 0.6871$\pm$0.0488 & 0.0943$\pm$0.0115 \\
        \hline
        \textbf{SemiGNN} & 0.5180$\pm$0.1358 & 0.1105$\pm$0.0794 & 0.5047$\pm$0.0934 & 0.2588$\pm$0.2012 & 0.4824$\pm$0.0669 & 0.0550$\pm$0.0076 \\
        \textbf{GDN} & 0.7769$\pm$0.0351 & 0.1752$\pm$0.0288 & 0.8017$\pm$0.0070 & 0.1789$\pm$0.0402 & 0.5798$\pm$0.0289 & 0.0680$\pm$0.0050 \\
        \textbf{LHML} & 0.7359$\pm$0.0146 & 0.2052$\pm$0.0259 & 0.8140$\pm$0.0130 & 0.1739$\pm$0.0518 & 0.7198$\pm$0.0063 & 0.0991$\pm$0.0037 \\
        \textbf{GCN-AugAN} & 0.6881$\pm$0.0315 & 0.1881$\pm$0.0059 & 0.8311$\pm$0.0390 & 0.2358$\pm$0.0209 & 0.7490$\pm$0.0213 & 0.1171$\pm$0.0050 \\
        \textbf{GDN-AugAN} & 0.8063$\pm$0.0301 & 0.2780$\pm$0.0517 & 0.8079$\pm$0.0218 & 0.2091$\pm$0.0295 & 0.7577$\pm$0.0317 & 0.1187$\pm$0.0092 \\
        \textbf{{\m}} & \textbf{0.8739$\pm$0.0105} & \textbf{0.3245$\pm$0.0159} & \textbf{0.8507$\pm$0.0146} & \textbf{0.4868$\pm$0.0311} & \textbf{0.7902$\pm$0.0045} & \textbf{0.1613$\pm$0.0103} \\
    \hline
    \end{tabular}\label{tab:main results injected}
\end{table*}

\begin{table*}[!t]
    \centering
    \caption{Few-shot performance on the datasets with synthetic anomalies. Best results are shown in bold.}
    \begin{tabular}{l|cc|cc|cc}
    \hline
        \multirow{2}{*}{\textbf{Methods}} & \multicolumn{2}{c|}{\textbf{Cora}} & \multicolumn{2}{c|}{\textbf{Citeseer}}&\multicolumn{2}{c}{\textbf{Amazon Photo}} \\ \cline{2-7}
        & AUC-ROC & AUC-PR & AUC-ROC & AUC-PR & AUC-ROC & AUC-PR\\ \hline \hline
        \textbf{1-shot} & 0.8662$\pm$0.0046 & 0.3092$\pm$0.0351 & 0.8445$\pm$0.0134 & 0.4555$\pm$0.0666 & 0.7809$\pm$0.0044 & 0.1647$\pm$0.0205 \\
        \textbf{3-shot} & \textbf{0.8750$\pm$0.0111} & 0.3237$\pm$0.0281 & 0.8435$\pm$0.0151 & 0.4283$\pm$0.0652 & 0.7892$\pm$0.0005 & 0.1455$\pm$0.0048 \\
        \textbf{5-shot} & 0.8747$\pm$0.0118 & 0.3232$\pm$0.0299 & 0.8415$\pm$0.0155 & 0.4219$\pm$0.0768 & 0.7892$\pm$0.0035 & \textbf{0.1648$\pm$0.0045} \\
        \textbf{10-shot} & 0.8739$\pm$0.0105 & \textbf{0.3245$\pm$0.0159} & \textbf{0.8507$\pm$0.0146} & \textbf{0.4868$\pm$0.0311} & \textbf{0.7902$\pm$0.0045} & 0.1613$\pm$0.0103 \\
    \hline
    \end{tabular}\label{tab:few shot injected}
\end{table*}

\subsection{Experimental Settings.} 
We introduce the experimental settings in details, including the compared baseline methods and evaluation metrics.\\
\textbf{Baseline Methods.}
We compare the {\m} with the following three categories of graph anomaly detection methods, including (1) shallow methods (i.e., \textbf{Radar}~\cite{li2017radar} and \textbf{ANOMALOUS}~\cite{peng2018anomalous}), 
(2) unsupervised deep learning methods (i.e., \textbf{GAAN}~\cite{chen2020generative}, \textbf{OCGNN}~\cite{wang2021one}, \textbf{DOMINANT}~\cite{ding2019deep}, \textbf{CoLA}~\cite{liu2021anomaly}, \textbf{ANEMONE}~\cite{jin2021anemone} and \textbf{CONAD}~\cite{xu2022contrastive}), and (3) semi-supervised deep learning methods (i.e., \textbf{Semi-GNN}~\cite{wang2019semi}, \textbf{GDN}~\cite{ding2021few}, \textbf{LHML}~\cite{guo2022learning}, \textbf{GCN-AugAN}~\cite{zhou2023improving} and \textbf{GDN-AugAN}~\cite{zhou2023improving}). Note that other semi-supervised methods are out of scope of this paper. For example, Meta-GDN~\cite{ding2021few} is focused on multiple graphs, and SAD~\cite{tian2023sad} is specifically designed over dynamic grpah.
\\
\textbf{Evaluation Metrics.}
\textbf{AUC-ROC and AUC-PR} are two widely used metrics in the existing literature~\cite{liu2022bond} where the classes are imbalanced.
\\\textbf{Implementation Details.}
In the experiments, we split the nodes into 80\% for training, 10\% for validation and 10\% for testing because the anomaly detector needs more data while RAN needs a relative small number of data as discussed in Section~\ref{sec:framework}. We set the number of few-shot labeled anomalies as 10. The dimension of representations is set as 64. For fair comparison, we use PyGOD\footnote{https://github.com/pygod-team/pygod}~\cite{liu2022pygod, liu2022bond} to implement unsupervised learning baseline methods because it provides a unified interface to implement the majority of unsupervised graph anomaly detection algorithms. We run all experiments multiple times and get the average and standard deviation.

\subsection{Graph Anomaly Detection Results (RQ1).}
\textbf{Overall Comparison.} We evaluate {\m} and the baseline methods in terms of AUC-ROC and AUC-PR. We show the performance comparison in Table~\ref{tab:main results injected} and Table~\ref{tab:main results organic} on the datasets with synthetic anomalies and with real anomalies, respectively. Accordingly, we have observations as follows:
\begin{itemize}
    \item {\m} can significantly outperform the existing graph anomaly detection methods, including shallow methods, unsupervised deep learning methods, and semi-supervised deep learning methods, on all datasets. For example, on YelpChi dataset, our method {\m} outperforms GDN by $13.76\%$ w.r.t. AUC-ROC. The major reason is that {\m} is able to fully exploit few labeled anomalies, and successfully adapt the knowledge from self-supervised learning to facilitate few-shot supervised learning with the proposed meta-learning algorithm.
    \item Shallow methods do not achieve desirable performance as they cannot utilize powerful expressive ability of GNNs to extract graph information. Unsupervised deep learning methods do not outperform semi-supervised deep learning methods, as they cannot utilize label information. Other semi-supervised deep learning methods, i.e., SemiGNN and GDN, fail to deliver satisfying performance compared to {\m}. The reason is SemiGNN needs to access multi-view labeled data and GDN is sensitive to standard deviation of the assumed Gaussian distribution.
\end{itemize}

\begin{table*}[!t]
    \centering
    \caption{Performance comparison results on the datasets with organic anomalies. Best results are shown in bold. OOM indicates out of memory.}
    \begin{tabular}{l|cc|cc|cc}
    \hline
        \multirow{2}{*}{\textbf{Methods}} & \multicolumn{2}{c|}{\textbf{Wiki}} & \multicolumn{2}{c|}{\textbf{Amazon Review}}&\multicolumn{2}{c}{\textbf{YelpChi}} \\ \cline{2-7}
        & AUC-ROC & AUC-PR & AUC-ROC & AUC-PR & AUC-ROC & AUC-PR\\ \hline \hline
        \textbf{Radar} & 0.4851$\pm$0.0000 & 0.0237$\pm$0.0000 & 0.1938$\pm$0.0000 & 0.0644$\pm$0.0000 & 0.2631$\pm$0.0000 & 0.0321$\pm$0.0000 \\
        \textbf{ANOMALOUS} & 0.5018$\pm$0.0085 & 0.0234$\pm$0.0003 & 0.1938$\pm$0.0000 & 0.0644$\pm$0.0000 & 0.2647$\pm$0.0032 & 0.0350$\pm$0.0034 \\ \hline
        \textbf{GAAN} & 0.4982$\pm$0.0572 & 0.0244$\pm$0.0035 & 0.2870$\pm$0.0445 & 0.0450$\pm$0.0024 & 0.4976$\pm$0.0197 & 0.0512$\pm$0.0023 \\
        \textbf{OCGNN} & 0.4579$\pm$0.0396 & 0.0204$\pm$0.0010 & OOM & OOM & 0.5218$\pm$0.0502 & 0.0636$\pm$0.0072 \\
        \textbf{DOMINANT} & 0.4401$\pm$0.0273 & 0.0198$\pm$0.0014 & 0.2698$\pm$0.0003 & 0.0433$\pm$0.0000 & 0.4895$\pm$0.0699 & 0.0597$\pm$0.0153 \\
        \textbf{CoLA} & 0.5435$\pm$0.0690 & 0.0263$\pm$0.0050 & 0.4573$\pm$0.1829 & 0.0933$\pm$0.0524 & 0.3379$\pm$0.0705 & 0.0479$\pm$0.0177 \\
        \textbf{ANEMONE} & 0.4402$\pm$0.0318 & 0.0203$\pm$0.0006 & 0.4576$\pm$0.0887 & 0.0816$\pm$0.0284 & 0.3359$\pm$0.0120 & 0.0362$\pm$0.0015 \\
        \textbf{CONAD} & 0.4396$\pm$0.0353 & 0.0198$\pm$0.0018 & 0.2698$\pm$0.0003 & 0.0433$\pm$0.0000 & 0.4873$\pm$0.0774 & 0.0600$\pm$0.0159 \\
        \hline
        \textbf{SemiGNN} & 0.4784$\pm$0.1270 & 0.0403$\pm$0.0147 & OOM & OOM & OOM & OOM \\
        \textbf{GDN} & 0.5096$\pm$0.0288 & 0.0229$\pm$0.0006 & 0.8016$\pm$0.0105 & 0.2761$\pm$0.0071 & 0.7131$\pm$0.0517 & 0.1755$\pm$0.0445 \\
        \textbf{LHML} & 0.6176$\pm$0.0310 & 0.0417$\pm$0.0029 & 0.7865$\pm$0.0092 & 0.3385$\pm$0.0138 & 0.7510$\pm$0.0484 & 0.2557$\pm$0.0290 \\
        \textbf{GCN-AugAN} & 0.6132$\pm$0.0175 & 0.0219$\pm$0.0010 & 0.8117$\pm$0.0275 & 0.3518$\pm$0.0232 & 0.7373$\pm$0.0178 & 0.2461$\pm$0.0190 \\
        \textbf{GDN-AugAN} & 0.6500$\pm$0.0293 & 0.0303$\pm$0.0279 & 0.8199$\pm$0.0388 & 0.3619$\pm$0.0187 & 0.7961$\pm$0.0296 & 0.2719$\pm$0.0033 \\
        \textbf{{\m}} & \textbf{0.6783$\pm$0.0197} & \textbf{0.0426$\pm$0.0031} & \textbf{0.8334$\pm$0.0091} & \textbf{0.4096$\pm$0.0422} & \textbf{0.8112$\pm$0.0031} & \textbf{0.3160$\pm$0.0025} \\
    \hline
    \end{tabular}\label{tab:main results organic}
\end{table*}

\begin{table*}[!t]
    \centering
    \caption{Few-shot performance on the datasets with organic anomalies. Best results are shown in bold.}
    \begin{tabular}{l|cc|cc|cc}
    \hline
        \multirow{2}{*}{\textbf{Methods}} & \multicolumn{2}{c|}{\textbf{Wiki}} & \multicolumn{2}{c|}{\textbf{Amazon Review}}&\multicolumn{2}{c}{\textbf{YelpChi}} \\ \cline{2-7}
        & AUC-ROC & AUC-PR & AUC-ROC & AUC-PR & AUC-ROC & AUC-PR\\ \hline \hline
        \textbf{1-shot} & 0.6776$\pm$0.0200  & 0.0425$\pm$0.0031 & 0.8296$\pm$0.0078 & 0.4079$\pm$0.0384 & \textbf{0.8124$\pm$0.0063} & 0.3136$\pm$0.0061 \\
        \textbf{3-shot} & 0.6777$\pm$0.0200 & 0.0425$\pm$0.0032 & 0.8302$\pm$0.0065 & 0.4153$\pm$0.0434 & 0.8105$\pm$0.0045 & 0.3136$\pm$0.0065 \\
        \textbf{5-shot} & 0.6778$\pm$0.0198 & 0.0426$\pm$0.0031 & 0.8320$\pm$0.0094 & \textbf{0.4204$\pm$0.0446} & 0.8116$\pm$0.0035 & 0.3119$\pm$0.0077 \\
        \textbf{10-shot} & \textbf{0.6783$\pm$0.0197} & \textbf{0.0426$\pm$0.0031} & \textbf{0.8334$\pm$0.0091} & 0.4096$\pm$0.0422 & 0.8112$\pm$0.0031 & \textbf{0.3160$\pm$0.0025} \\
    \hline
    \end{tabular}\label{tab:few shot organic}
\end{table*}

\textbf{Few-Shot Evaluation.} To verify the effectiveness of {\m} in the few-shot, we evaluate {\m} by providing different numbers of labeled anomalies. The results are summarized in Table~\ref{tab:few shot injected} and Table~\ref{tab:few shot organic}. According to the results, we have the following observations:
\begin{itemize}
    \item {\m} can obtain desirable performance and outperform the baselines in the Table~\ref{tab:main results organic} even if only 1-shot anomaly is provided. For example, on Wiki dataset, {\m} with 1-shot anomaly outperforms SemiGNN by $32.97\%$ w.r.t. AUC-ROC. It confirms the effectiveness of {\m} to utilize the very limited labeled anomalies.
    \item The performance of {\m} does not strictly increase with the number of labelled anomalies due to the complexity and irregularity of anomalies. Combing the previous observations, we conclude that it is important to incorporate anomalies for training models, while the performance of {\m} is not sensitive to the number of anomalies.
\end{itemize}

\subsection{Ablation Study (RQ2).}
To verify the effectiveness of each key component of {\m}, we design an ablation study on the following variants of MetaGAD:
\begin{itemize}
    \item \textbf{\textsc{FinetuneGAD}} adopts the pretrain-finetune method. It conducts finetuning with few-shot anomalies on the embedding from the graph encoder.
    \item $\textbf{\textsc{MetaGAD}}^-$ excludes RAN compared to {\m}. The representation from the graph encoder is directly followed by the anomaly detector.
    \item \textbf{\textsc{JointGAD}} jointly trains a randomly initialized graph encoder and other modules.
    \item \textbf{\textsc{JointPretrainGAD}} jointly trains a pretrained graph encoder and other modules.
\end{itemize}

\begin{figure}[!t]
    \centering
    \includegraphics[width=0.5\textwidth]{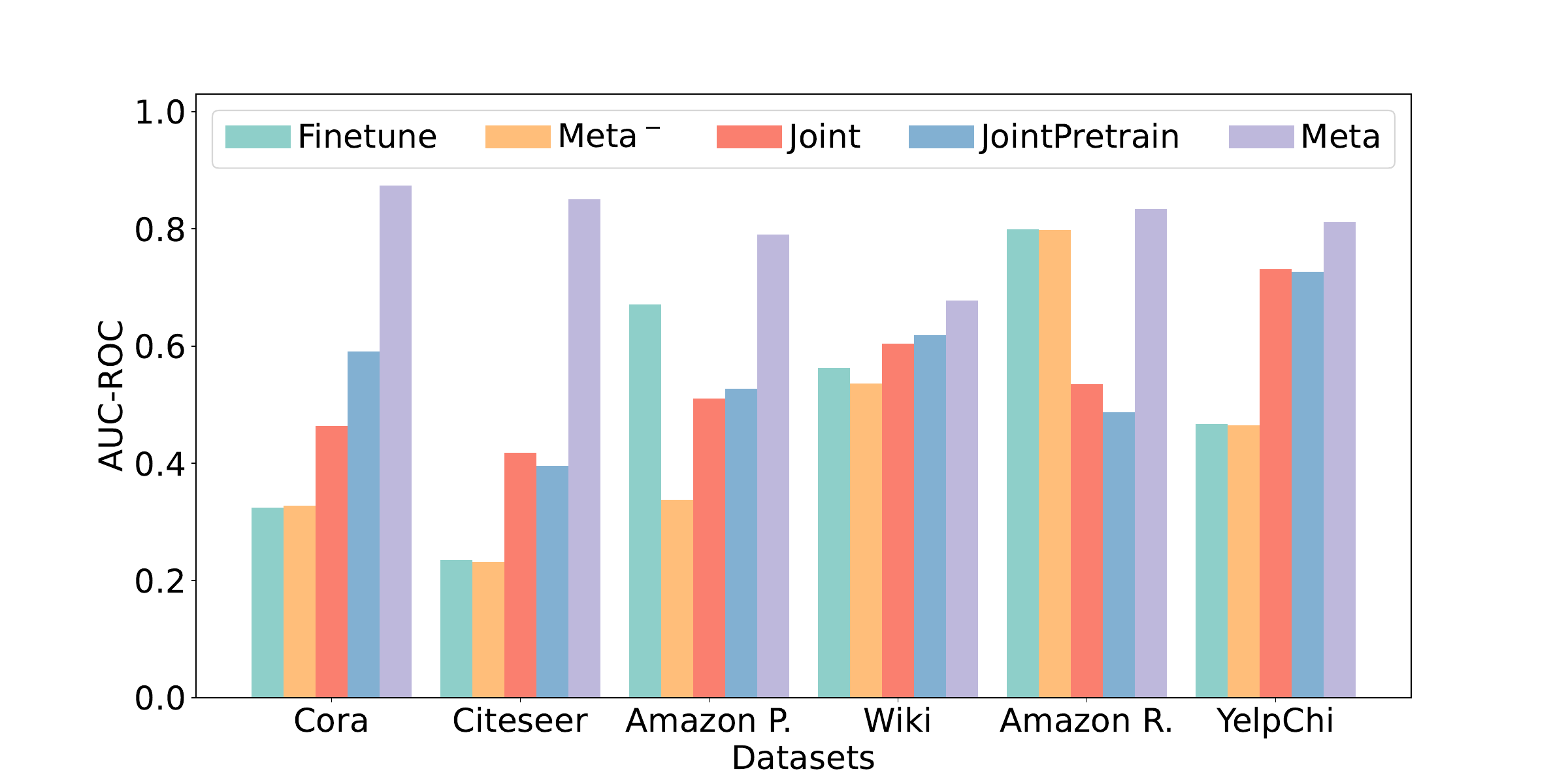}
     \caption{Ablation study of {\m} w.r.t. AUC-ROC on all datasets. Amazon P. and Amazon R. are abbreviations of Amazon Photo and Amazon Review, respectively.}
     \vspace{-0.5cm}
    \label{fig:ablation}
\end{figure}

We abbreviate the above five methods as \textsc{Finetune}, $\textsc{Meta}^-$,  \textsc{Joint},  \textsc{JointPretrain} and \textsc{Meta}, respectively. From the experimental results shown in Figure~\ref{fig:ablation}, we observe the following points :
\begin{itemize}
    \item Compared to \textsc{Finetune} and $\textsc{Meta}^-$, \textsc{Meta} obtains desirable performance. The primary reason is that upon the proposed meta-learning algorithm, RAN can adapt the raw representation to the anomaly-aware representation and largely alleviate the overfitting issue, with incorporating validation loss feedback. It empirically verifies the effectiveness and the necessity of the meta-learning algorithm and RAN.
    \item Compared to \textsc{Joint} and \textsc{JointPretrain}, \textsc{Meta} achieves impressive performance. The major reason is that jointly training the graph encoder with other modules introduces interference. The graph encoder focus on extracting structure and feature information of the graph, whereas the other modules aim to adapt information from the graph encoder to facilitate few-shot supervised learning. It demonstrates the reasonable design of \textsc{Meta}.
\end{itemize}

\vspace{-0.3cm}
\subsection{Overfitting Issue Investigation (RQ3).}
\begin{figure}[!t]
    \centering
    \includegraphics[width=0.5\textwidth]{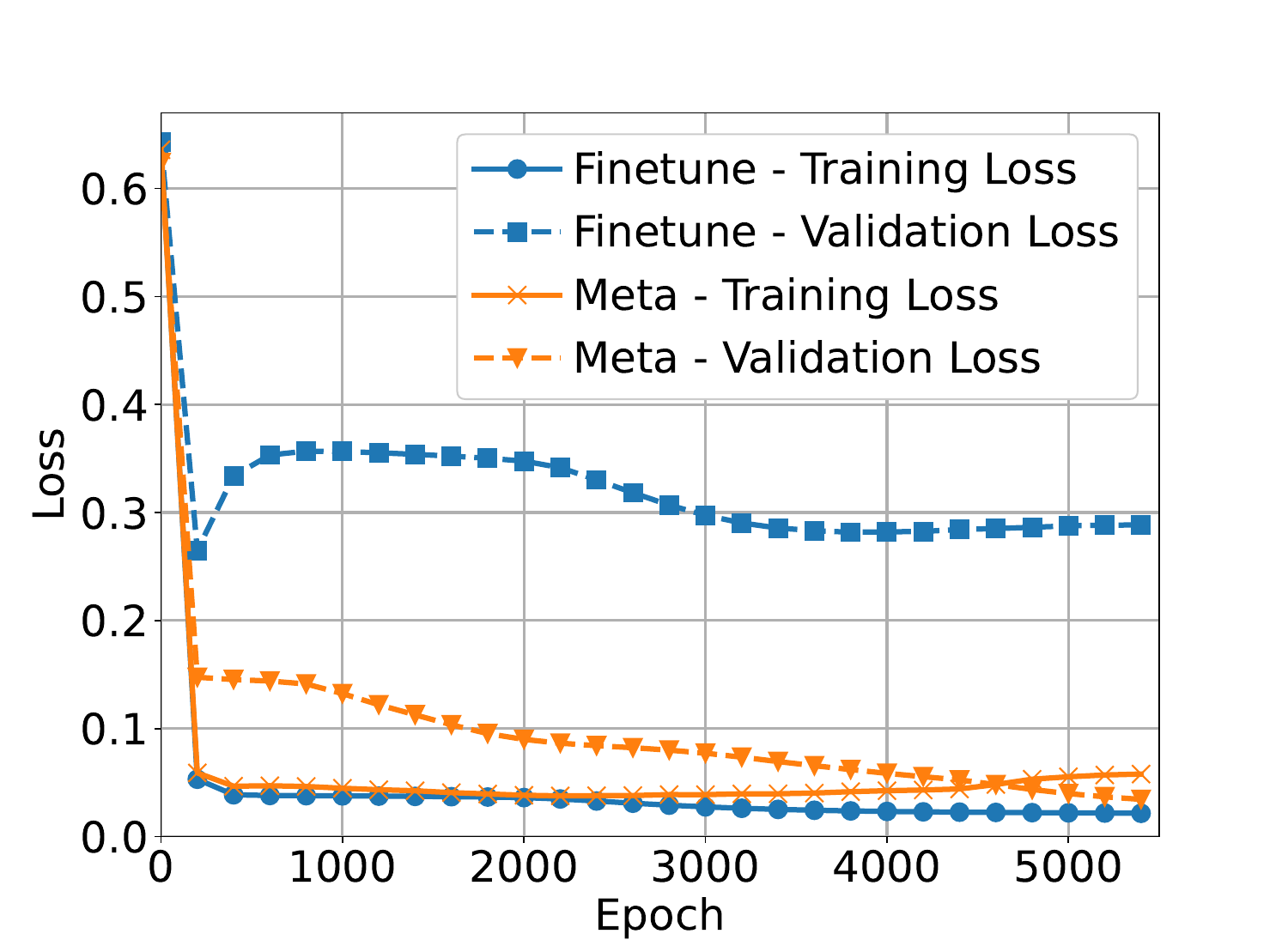}
     \caption{The training and validation loss for \textsc{Finetune} and \textsc{Meta}. \textsc{Finetune} falles into the overfitting problem while \textsc{Meta} largely alleviates the issue.}
     \vspace{-0.5cm}
    \label{fig:loss}
\end{figure}

To verify if the {\m} can deal with the notorious overfitting issue in the few-shot learning, we show how the training loss and validation loss for \textsc{Meta} ({\m}) and \textsc{Finetune} (\textsc{FinetuneGAD}) change with epochs in Figure~\ref{fig:loss}. We observe that for \textsc{Finetune}, the training loss decreases with epochs, but the validation loss first decreases to the minimum value at around 200th epochs then starts increasing. It is clear \textsc{Finetune} falls into the overfitting issue. For \textsc{Meta}, although the training loss first decreases then gradually increases, the validation loss keeps decreasing with epochs throughout the entire training process. It shows \textsc{Meta} can largely alleviate the overfitting issue. The reason is that in \textsc{Meta}, the validation loss, as an indicator of generalization performance, is incorporated into the optimization process, thus ensuring the optimization direction is towards minimizing the validation loss.

Note that \textsc{Meta} does not have information leakage problem. In \textsc{Finetune}, the model is trained with the training set and then hyperparameter tuning is performed based on the loss on the validation set. In \textsc{Meta}, the model is trained with the training set and the loss on the validation set is incorporated to guide the optimization process. Both \textsc{Meta} and \textsc{Finetune} utilize the same information (loss) of the validation set. The difference is that \textsc{Meta} is capable of leveraging the validation loss to guide the training rather than just for hyperparameter tuning. So \textsc{Meta} has better capacity in terms of incorporation the limited labeled anomalies and thus improve the anomaly detection performance significantly.

\subsection{Class Imbalance Investigation (RQ4).}\label{sec:imbalance}
\begin{figure*}[th]
    \centering
    \vspace{-0.3cm}
    \subfigure[AUC-ROC]{
        \begin{minipage}{0.49\textwidth}
            \includegraphics[width=1\textwidth]{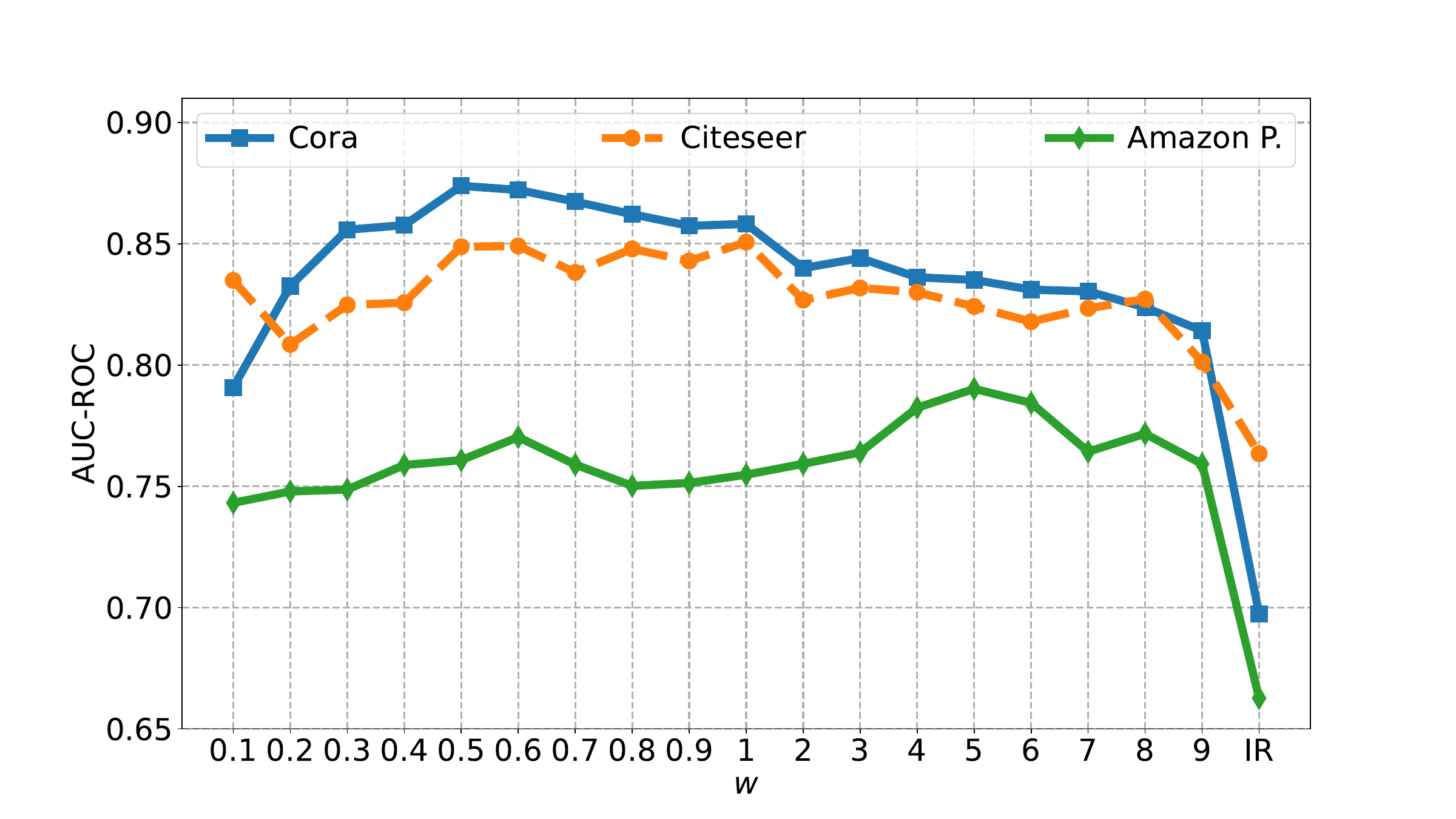}
        \end{minipage}\label{fig:imbalanceauc}
    }
    \hspace{-0.5cm}
    \subfigure[AUC-PR]{
        \begin{minipage}{0.49\textwidth}
            \includegraphics[width=1\textwidth]{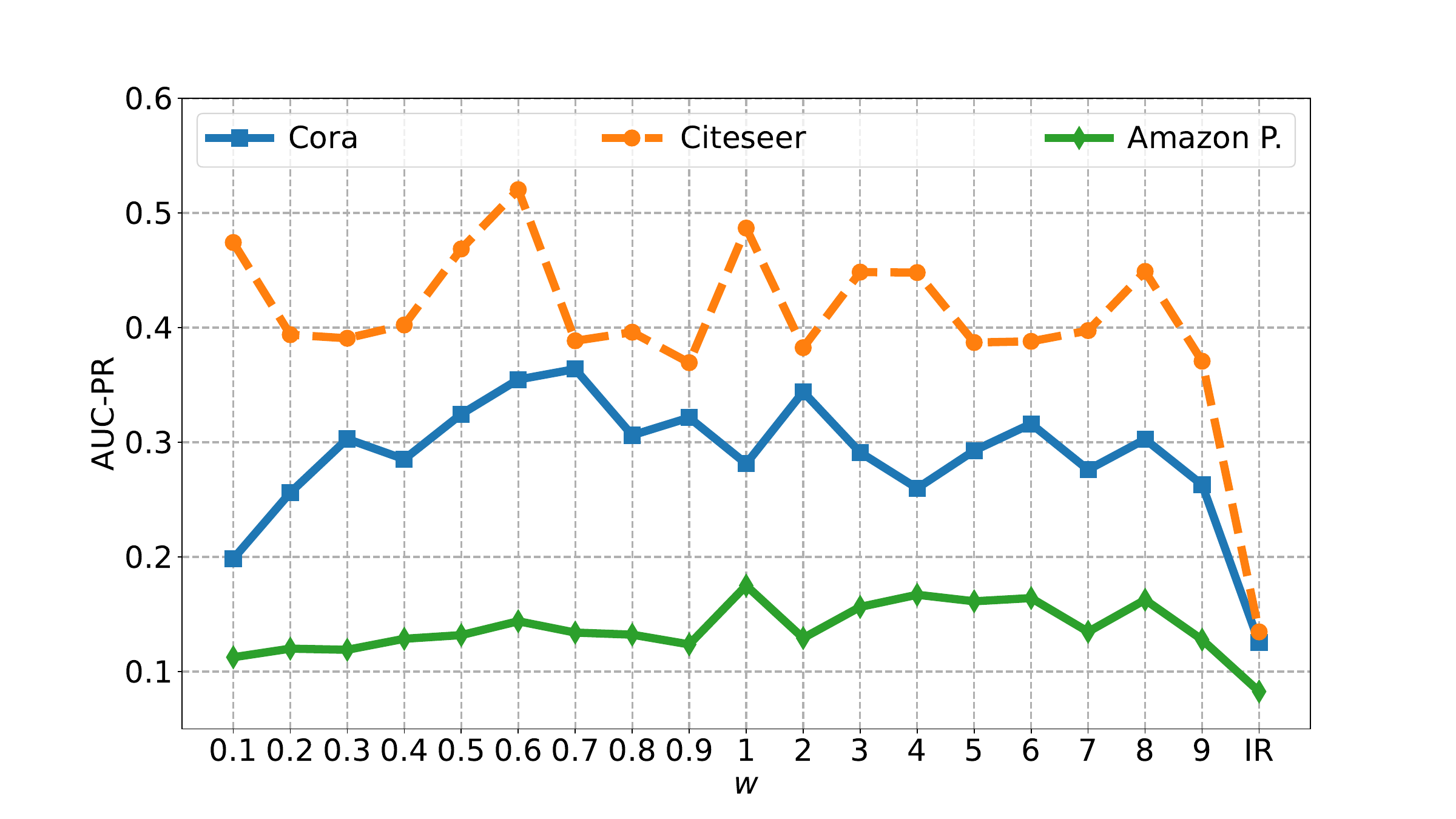}
        \end{minipage}\label{fig:imbalancepr}
    }
    \vspace{-0.3cm}
    \caption{Effect of cost weight $w$ on the performance of {\m} on three dataset with synthetic anomalies w.r.t. (a) AUC-ROC and (b) AUC-PR. The IR values of Cora, Citeseer and Amazon Photo are 216, 235 and 611, respectively. However, the optimal value of $w$ w.r.t. AUC-ROC (a) are 0.5, 1 and 5, respectively; the optimal value of $w$ w.r.t. AUC-PR (b) are 0.7, 0.6 and 1, respectively.}
    \label{fig:imbalance}
    \vspace{-0.5cm}
\end{figure*}

In this part, we investigate the effect of cost weight $w$ in the loss function on performance of {\m}. We define the $IR$ (Imbalance Ratio) value as $IR = \frac{|positive\ instances|}{|negative\ instances|}$. When the cost weight $w$ is equal to $IR$, the training is considered balanced as the large cost weight value compensates the disparity between the number of positive instances and negative instances. By providing different weight values $w$ for {\m}, we have the performance trends as shown in Figure~\ref{fig:imbalance}. We find an interesting fact that the imbalance between positive instances and negative instances does not hurt the anomaly detection performance; instead, the imbalance in a degree is the better choice for the performance. For example, from the Figure~\ref{fig:imbalanceauc}, we observe that the optimal $w$ for {\m} on Cora, Citeseer, and Amazon Photo w.r.t. AUC-ROC are 0.5, 1 and 5, respectively; similarly, from the Figure~\ref{fig:imbalancepr}, we notice that the optimal $w$ for {\m} on Cora, Citeseer, and Amazon Photo w.r.t. AUC-PR are  0.7, 0.6 and 1, respectively. The above optimal values of $w$ are apparently less than $IR$ values. The reason is that the graph anomaly detection problem itself is a highly-imbalanced problem where the number of normal instances is way too more than that of anomalies in real-world scenarios, e.g., the number of normal accounts is much more than that of suspicious accounts in the bank. Therefore, the anomaly detector in the training should be provided with similar imbalanced distribution (i.e., highly skewed distribution) with the test data (real-world scenarios).

\subsection{Contamination Level Investigation (RQ5)}\label{sec:contamination}
\begin{figure}[th!]
    \centering
    \vspace{-0.2cm}
    \hspace{-0.3cm}
    \subfigure[Effect of CR]{
    	\begin{minipage}{0.24\textwidth}
   		 	\includegraphics[width=\textwidth]{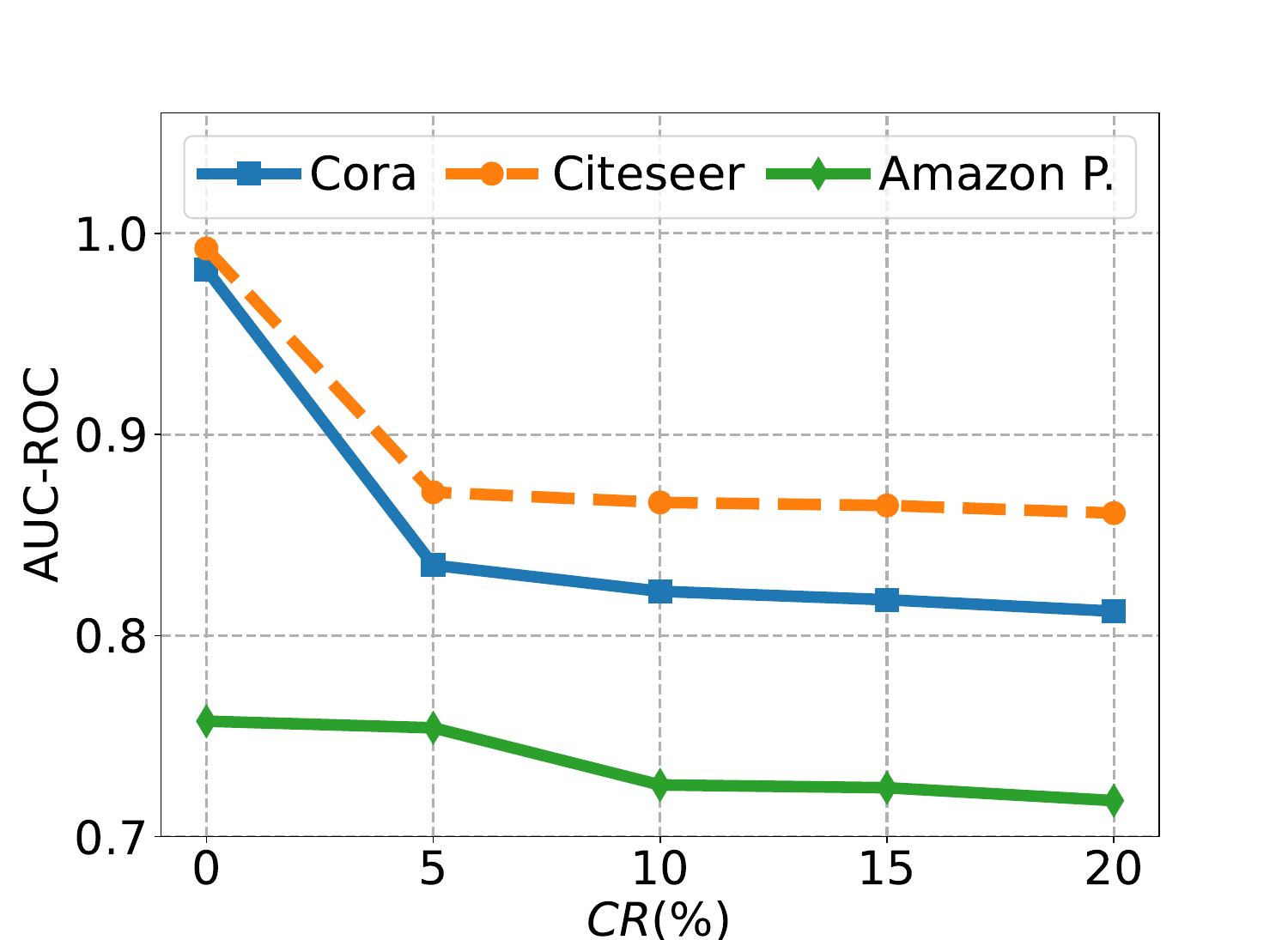}
    	\end{minipage}\label{fig:contaminationa}
    }
    \hspace{-0.4cm}
    \subfigure[The statistics of $m$ and $n$]{
    	\begin{minipage}{0.24\textwidth}
   		 	\includegraphics[width=\textwidth]{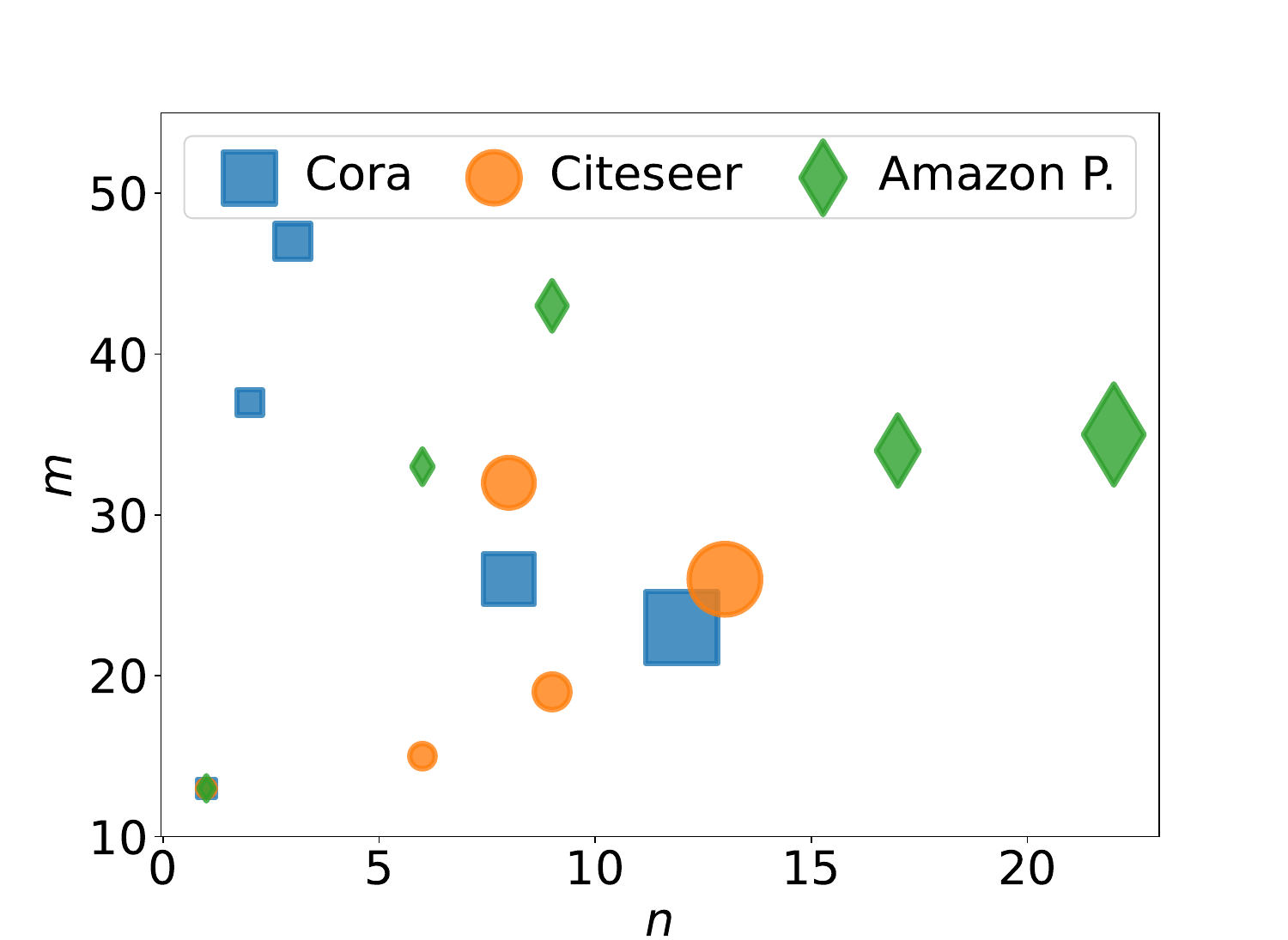}
    	\end{minipage}\label{fig:contaminationb}
    }
    \caption{(a) Effect of contamination levels on performance of {\m}. (b) The statistics of $m$ and $n$. The marker with five sizes for a dataset in (b) corresponds to five CR values in (a), from small to large, respectively. For example, $n=9$ and $m=43$ for Amazon P. corresponds to $CR=10\%$; $n=22$ and $m=35$ for Amazon P. corresponds to $CR=20\%$. Note that $CR=0\%$ corresponds to $n=1$ and $m=13$ for all datasets.}
    \label{fig:contamination}
    \vspace{-0.7cm}
\end{figure}

In this subsection, we analyze the effect of {\m} at different levels of contamination. As discussed in Section~\ref{sec:framework}, since all unlabeled nodes are regarded as normal nodes in the training, such simple training strategy introduces $CR$ (Contamination Ratio) into the training data. Formally, we define the $CR$ as $CR = \frac{|unlabeled\ anomalies|}{|unlabeled\ nodes|}.$
To generate datasets with different contamination levels, we inject different number of anomalies into Cora, Citeseer and Amazon Photo. Specifically, we set different numbers of $m$ and $n$ for structural anomalies described in ~\ref{sec:dataset}, and synthesize contextual anomalies with the same number of structural anomalies. We provide {\m} with datasets with different $CR$ values. The evaluation results are shown in Figure~\ref{fig:contaminationa} and the scatter plot of $m$ and $n$ is depicted in Figure~\ref{fig:contaminationb}. From the figures, we have the following observations:
\begin{itemize}
    \item The {\m} can obtain incredible performance when the $CR$ is equal to 0, where there is no any contamination in the training data. For example, on Cora and Citeseer datasets, the values of AUC-ROC are 0.9821 and 0.9925, respectively. The reason is that {\m} can be trained more efficiently without contamination in the training data.
    \item When the $CR$ is greater than 0, the contamination starts appearing in the training data. Even if the $n$ and $m$ vary greatly to synthesize different number of anomalies, the performance of {\m} decreases slightly with the increasing contamination level. For example, on Cora dataset, when the $CR$ is from $5\%$ to $20\%$, with $n$ varying from 2 to 12 and $m$ varying from 37 to 23, the performance of {\m} only decreases $2.74\%$. It demonstrates the {\m} is robust to noise in the training data.
\end{itemize}

\section{Conclusion and Future Work}\label{sec:conclusion}
\vspace{-0.2cm}
In this paper, we study an important and practical problem of few-shot graph anomaly detection. To solve the problem, we propose a novel meta-learning framework {\m} that learns to adapt the node representations from self-supervised learning to maximally facilitate few-shot supervised learning. {\m} can bridge the representation gap and largely alleviate the overfitting issue for the few-shot learning. Within {\m}, RAN acts as the meta-learner, adjusting node representations from self-supervised learning to better suit few-shot supervised learning. The anomaly detector, serving as the target model, accurately identifies anomalies based on the adapted representation. The meta-leaning algorithm enables RAN and the anomaly detector enhance each other synergistically where a bi-level optimization structure ensures {\m} avoids the notorious overfitting issue. The extensive experiments on multiple datasets with synthetic anomalies and real anomalies verify the effectiveness of {\m}.

Our work gives a new opportunity in various aspects. For research problem, the potential future work includes few-shot anomaly detection on more complex networks, e.g., heterogeneous graphs, dynamic graphs and hypergraphs, or other kinds of data, e.g., tabular data and image data. For technical methods, the meta-learning and the RAN can be applied in many problems where the representation gap exists. Moreover, exploring graph encoders from other self-supervised learning me, e.g., contrastive models, is an interesting direction.

\section*{Acknowledgments}
This material is based upon work supported by the U.S. Department of Homeland Security under Grant Award Number 17STQAC00001-07-04, NSF awards (SaTC-2241068, IIS-2339198, and POSE-2346158), a Cisco Research Award, and a Microsoft Accelerate Foundation Models Research Award. The views and conclusions contained in this document are those of the authors and should not be interpreted as necessarily representing the official policies, either expressed or implied, of the U.S. Department of Homeland Security.

\bibliographystyle{IEEEtran}
\bibliography{dsaa}

\end{document}